\documentclass[letterpaper, 10 pt, conference]{ieeeconf}  

\usepackage{algorithm}
\usepackage{algorithmic}

\usepackage{times}
\usepackage{helvet}
\usepackage{courier}
\usepackage{hyperref}
\usepackage{graphicx}
\usepackage[usenames,dvipsnames]{xcolor}
\usepackage{tabularx}
\usepackage{makecell}
\usepackage{subcaption}
\usepackage{amsfonts} 

\usepackage[%
    style=ieee,
    sorting=none,
    natbib=true,  
    backend=biber,
    sortcites=true,
    doi=false,
    url=false,
    mincitenames=1,
    maxcitenames=1, 
    minbibnames=1, 
    maxbibnames=4, 
    hyperref
]{biblatex}

\let\eqref\undefined

\newcommand{\figref}[1]{Figure~\ref{fig:#1}}
\newcommand{\algoref}[1]{Algorithm~\ref{algo:#1}}
\newcommand{\secref}[1]{Section~\ref{sec:#1}}
\newcommand{\tabref}[1]{Table~\ref{tab:#1}}
\newcommand{\eqref}[1]{Equation~\ref{eq:#1}}

\newcommand{\figlabel}[1]{\label{fig:#1}}
\newcommand{\algolabel}[1]{\label{algo:#1}}
\newcommand{\seclabel}[1]{\label{sec:#1}}
\newcommand{\tablabel}[1]{\label{tab:#1}}

\IEEEoverridecommandlockouts                              

\title{\LARGE \bf
Propagating Semantic Labels in Video Data
}

\author{David Balaban, Justin Medich, Pranay Gosar, and Justin Hart$^{1}$
\thanks{$^{1}$Departent of Computer Science, The University of Texas at Austin, Austin, Texas, USA 78712
        {\tt\small \{dbalaban, jmedich, pgosar, hart\}@cs.utexas.edu}}%
}

\begin{document}

\maketitle
\thispagestyle{empty}
\pagestyle{empty}

\begin{abstract}


Semantic Segmentation combines two sub-tasks: the identification of pixel-level image masks and the application of semantic labels to those masks. Recently, so-called Foundation Models have been introduced; general models trained on very large datasets which can be specialized and applied to more specific tasks. One such model, the Segment Anything Model (SAM), performs image segmentation. Semantic segmentation systems such as CLIPSeg and MaskRCNN are trained on datasets of paired segments and semantic labels. Manual labeling of custom data, however, is time-consuming. This work presents a method for performing segmentation for objects in video. Once an object has been found in a frame of video, the segment can then be propagated to future frames; thus reducing manual annotation effort. The method works by combining SAM with Structure from Motion (SfM). The video input to the system is first reconstructed into 3D geometry using SfM. A frame of video is then segmented using SAM. Segments identified by SAM are then projected onto the the reconstructed 3D geometry. In subsequent video frames, the labeled 3D geometry is reprojected into the new perspective, allowing SAM to be invoked fewer times. System performance is evaluated, including the contributions of the SAM and SfM components. Performance is evaluated over three main metrics: computation time, mask IOU with manual labels, and the number of tracking losses. Results demonstrate that the system has substantial computation time improvements over human performance for tracking objects over video frames, but suffers in performance.

\end{abstract}

\section{INTRODUCTION}

This work presents a Semantic Label Propagation (SLP) system for labeling large sets of images derived from video; providing a dataset suitable for training or fine-tuning models for computer vision tasks with minimal manual labeling. The system takes as input video from a moving camera in a static scene and outputs a set of objects, each object corresponds to a set of binary masks for each frame it is identified in.

\begin{figure}
    \centering
    \includegraphics[width=\linewidth]{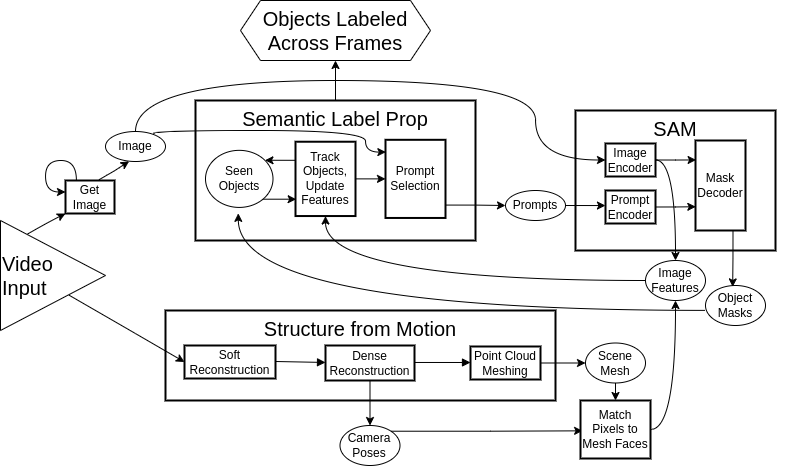}
    \caption{Semantic Label Propagation Overview}
    \label{SLPdiag}
\end{figure}

Figure \ref{SLPdiag} diagrams the flow of data through of the proposed system. The triangle on the left represents the ``Video Input'' to the system. The hexagon at the top represents the output, ``Objects Labeled Across Frames.'' Ellipses represent data, while squares represent sub-processes. This system begins with a Structure from Motion (SfM) pipeline to gain an object mesh~\cite{ozyecsil2017survey}. Then, Segment Anything Model (SAM) is used to identify the distinct objects in the mesh~\cite{kirillov2023segany}. These objects are then tracked through all frames of video utilizing the object mesh to enforce geometric consistency while SAM associates previously unseen polygons to tracked objects. 

To evaluate system performance, ground truth semantic labels are collected from five videos. Volunteers use the open-source PixelAnnotationTool to manually label objects in frame~\cite{Breheret:2017}. The volunteer work is spot checked by the authors. This data will be made publicly-available upon publication of this work, at the Texas Robotics Dataverse: \url{https://dataverse.tdl.org/dataverse/robotics}

To demonstrate the performance contribution of each component of the semantic label propagation pipeline, four variants are presented which vary how prompts are selected as input to SAM:

\begin{enumerate}
    \item SAM-only-1.0, $k$-best prompts
    \item SAM-only-2.0, hill-climb for best prompt
    \item SfM-SAM-1.0, $k$-best SfM candidate prompts
    \item SfM-SAM-2.0, $k$-random SfM candidate prompts
\end{enumerate}
These variants are fully described in Section \ref{sec:method}.

The novel contributions of this work are the use of SfM to propagate geometry into new frames, as described in \secref{SfMSAM}; enhancements of the use of feature embeddings provided by SAM, as described in \secref{SAM}; a direct comparison of automatic labeling of video to volunteer manual labeling, as described in \secref{eval}.

\section{Related Work}Tasks such as Object Detection~\cite{zaidi2022survey,terven2023comprehensive}, Image Captioning~\cite{stefanini2022show,li2022blip}, and Semantic Segmentation~\cite{mo2022review,li2022deep} are approaches to image labeling for the purposes of scene understanding~\cite{gan2022vision,awais2023foundational}. Object Detection localizes a number of desired objects within the image frame, typically with a bounding box for each object~\cite{zaidi2022survey,terven2023comprehensive}. Image Captioning gives a general description of the image, often using a text prefix provided to a large language model~\cite{stefanini2022show,li2022blip}. Semantic Segmentation provides pixel-level annotations of the image~\cite{mo2022review,li2022deep}.

Pre-trained models have been released for people to fine-tune with custom data~\cite{mokady2021clipcap,diwan2023object,Yakubovskiy:2019}, which can require an arduous amount of manual labeling. There has been a significant effort to design models that require as few training examples as possible~\cite{wang2020generalizing}. One such model is CLIP, which is a joint embedding between language and images that identifies the text which best describes an image~\cite{radford2021learning,mokady2021clipcap}. CLIP is pre-trained with multiple datasets, which recent work suggests leads to a lower quality of data~\cite{nguyen2022quality}. There is a need for a method which generates large high-quality datasets of semantically labeled images with minimal manual labeling.

Video Object Segmentation and Tracking (VOST) refers to the coupled nature of pixel-level annotations and object tracking, as solving one improves the solution to the other~\cite{yao2020video}. Multiple Object Tracking (MOT) identifies the trajectory many objects take through a scene~\cite{luo2021multiple}. 

Segment Anything Model (SAM) is a recently released vision transformer which effectively handles the annotation piece of the VOST problem~\cite{kirillov2023segany}. SAM segments images on a pixel-level without semantic annotations. With SAM, a single pixel manual label is sufficient to semantically label a whole object, and identify that object in other frames and contexts~\cite{zhang2023personalize}. This feature can be utilized to track objects across many frames of a video~\cite{yang2023track}.

Structure from Motion (SfM) is the problem of reconstructing a scene from images taken at different locations, such as in a video~\cite{ozyecsil2017survey}. Recent research has studied SfM methods in both outdoor and indoor settings~\cite{jiang2020efficient,pintore2020state}. Classic approaches to SfM utilize optimization schemes such as Levenberg-Marquardt~\cite{more2006levenberg} to jointly solve for camera poses and pixel depths, however more recent research uses a deep learning model for this optimization~\cite{wei2020deepsfm}.

Prior work on the collection of reliable datasets for training CV models with minimal manual labeling include utilizing 3D object representations jointly learned with object association to 2D image labels~\cite{he2023tracking}, and generating datasets by combining object segmentation with optical flow methods~\cite{porzi2020learning}.

\section{Label Propagation Procedures} \label{sec:method}

\algoref{SLP} describes the proposed system. The algorithm takes as input a SAM model, a video, a positive integer parameter $F$, and additional implementation-specific parameters described in \secref{FindObject}. A list of objects is tracked across video frames, objects in this list are referred to as known objects. Each known object has associated characteristic features of which there are two types: SAM features which describe an object's visual appearance, and mesh faces which describe an object's 3D geometry. One or both of these types may be used.

At each frame, known objects are identified in frame by searching for matching characteristic features. Searching for and updating characteristic features is handled by the FIND\_OBJECT function in \algoref{SLP}, described in detail in \secref{FindObject}. This function returns a binary mask identifying the object's location in frame, called the object mask. The object mask and frame is recorded as a successful label propagation.

Once all the known objects are handled, new objects are identified to add to the list of known objects. The union of all object masks, named the seen mask, is used to avoid searching for new objects in regions of the frame that have been associated to known objects. The parameter $F \in \mathbb{N}$ causes the procedure to skip $F$ frames before adding new objects for faster computation. All object masks are identified by SAM, as described in \secref{SAM}.

\begin{algorithm}
\caption{Semantic Label Propagation} \algolabel{SLP}
\begin{algorithmic}[1]
\REQUIRE SAM, Video, $F$, *args
\STATE Objs $\leftarrow \{\}$
\STATE $f \leftarrow 0$
\FOR{frame in Video}
    \STATE seen\_mask $\leftarrow$ empty
    \FOR{obj in Objs}
        \STATE obj\_mask $\leftarrow$ FIND\_OBJECT(frame, obj, *args)
        \IF{obj\_mask not empty}
            \STATE obj.append(frame, obj\_mask)
            \STATE seen\_mask $\leftarrow$ seen\_mask $\cup$ obj\_mask
        \ENDIF
    \ENDFOR
    \IF{$f \ge F$}
        \STATE seen\_mask $\leftarrow$ Closing(seen\_mask)
        \STATE new\_masks $\leftarrow$ SAM.getMasks(frame, seen\_mask)
        \STATE Objs.append(new\_masks, frame)
        \STATE $f \leftarrow 0$
    \ELSE
        \STATE $f \leftarrow f+1$
    \ENDIF
\ENDFOR
\RETURN Objs
\end{algorithmic}
\end{algorithm}

\subsection{Structure from Motion}
This paper uses a COLMAP to MVS implementation of SfM where COLMAP provides an initial guess to MVS~\cite{cernea2020openmvs,schoenberger2016sfm,schoenberger2016mvs}. Given a video input of a camera moving around a scene, COLMAP will extract distinctive features in the scene that can be matched between frames of the video. MVS enhances this process by allowing the SfM pipeline to estimate the depth of objects within the frames, allowing for a more detailed 3D model. The pipeline outputs a 3D reconstruction of the scene in the form of a mesh of polygon faces, called the scene mesh. It also outputs the video and camera poses that capture the estimated positions and orientations of the camera used. This output is sufficient to identify which mesh face every pixel observes. Segmented frames outputted by SAM are then projected onto the reconstruction which allows less invocations to SAM for future frames.

\subsection{Segment Anything Model}
SAM takes a set of $k\ge1$ pixel locations as a prompt, and returns a binary mask associated with each prompt to identify objects in frame. SAM identifies masks with a joint image and feature encoding. Image encoding provides a feature vector at each pixel location, while the prompt encoding provides a feature vector for each prompt. The cosine similarity between feature vectors indicates the likelihood of a prompt-pixel pair being a part of the same segment. 

\subsection{Getting Object Masks from SAM}
\seclabel{SAM}
When tracking known objects, a prompt is provided to SAM as a sample of the object's expected location in frame, producing an object mask. The union of all known object masks in the current frame is the seen mask.

SAM.getMasks() searches for unseen objects, this function uses a grid of $g$x$g$ single-pixel prompts, called the prompt grid. The seen mask is used to filter prompts that have already been associated to known objects, with $g'$, $0\leq g' \leq g^2$, total prompts used in each frame. SAM will then produce $M$, $0\leq M \leq g'$, new object masks.

The seen object mask tends to exclude borders between objects, but due to the proximity of known objects, prompts in these regions tend to produce duplicates or near-duplicates of these objects, leading to longer computation times. To counter this problem, the Closing image morphology is applied to the seen mask from the open-source software OpenCV, thereby adding the object boarders to the seen mask~\cite{opencv_library,serra1982image}.

OpenCV's Closing Image Morphology transforms a binary mask in an iterative process. When Closing, at each iteration two other morphologies are applied in succession, a dilation followed by an erosion. The dilation expands the white space of the binary mask and the erosion expands the black space. The resulting effect is that small regions of black space are removed while boundaries between substantially sized white and black regions are maintained. By applying this morphology to the seen mask, the boarders between known objects is eliminated while unseen regions remain unaltered.~\cite{opencv_library,serra1982image}

\figref{seenMask} shows the effect of Closing() with a color coded filter added to a frame of video. Blue regions show the raw seen mask, red shows the regions added to the seen mask by Closing(), and green shows the final unseen mask. Only the prompt grid points in the green regions (as seen in the top of Figure 2) will be used to find new objects, which significantly reduces the number of duplicated tracked objects without sacrificing the ability to observe new objects. In individual images, the green regions are small and appear only at the edges of individual frames. This leads to a substantial speed-up over running SAM on every frame of video.

\begin{figure}
    \centering
    \vspace{2mm}
    \includegraphics[width=\linewidth]{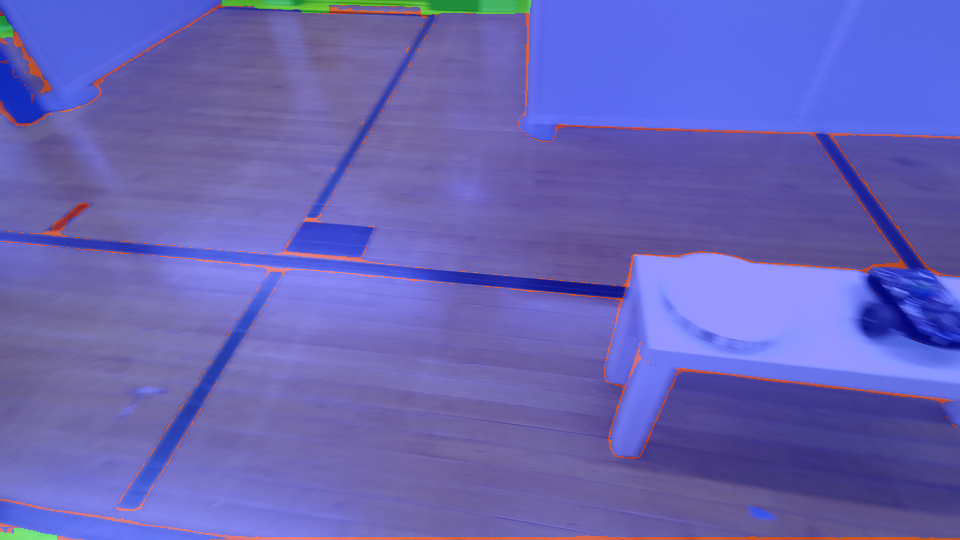}
    \caption{Seen mask image morphography by color filter: blue, raw seen mask; red, seen by image morphography; green, unseen.}
    \figlabel{seenMask}
\end{figure}

\subsection{FIND\_OBJECT Procedure}
\seclabel{FindObject}
FIND\_OBJECT() in \algoref{SLP} takes as input the current frame and a desired object to identify in frame and returns the object mask if found. It also handles the update of characteristic features for the known object: visual, geometric, or both. Additional parameters are required depending on which features are used, corresponding to different function implementations. 

\subsubsection{SAM Only Features}
\seclabel{SAMonly}
\algoref{SAMonly} describes the FIND\_OBJECT() implementation when only visual features are used, it takes an additional integer parameter $k$ as input. Visual features come from SAM's image and prompt joint encoding.  

The first time an object is identified, the prompt features associated with the object mask returned by SAM  are recorded as the characteristic visual feature. These objects can be identified in future frames by maximizing the cosine similarity of the recorded features across the image encoding. 

\subsubsection{K-best approach}
Pixels are scored by the cosine similarity, and the $k$-best pixels are used to generate a prompt that identifies the object in frame. When $k>0$, all pixels are scored, which can be an expensive process with large feature vectors.

\subsubsection{Hill-climbing approach}
When $k\leq0$, the last prompt location is recorded along with the visual features. This location is used as an initial guess to a discrete hill-climbing procedure that searches for the maximum score locally. At each step of the hill-climb, only the neighboring pixels need be scored, choosing the best from among them as the location for the next iteration until all neighbors score lower than the current location. This final location is used as the single-pixel prompt.

The feature vector is updated to account for changes in lighting and viewing angles.

\begin{algorithm}
\caption{SAM-only FIND\_OBJECT} \algolabel{SAMonly}
\begin{algorithmic}[1]
\REQUIRE frame, obj, SAM, $k$
\STATE image\_features $\leftarrow$ SAM.encodeImage(frame)
\STATE pixel\_scores $\leftarrow$ DotProd(obj.features, image\_features)
\IF{$k>0$}
    \STATE prompt $\leftarrow$ kBestScores(pixel\_scores)
\ELSE
    \STATE prompt $\leftarrow$ hillClimb(pixel\_scores, obj.last\_prompt)
\ENDIF
\IF{score at prompt $<$ threshold}
    \RETURN none
\ENDIF
\STATE obj.last\_prompt $\leftarrow$ prompt
\STATE obj.features $\leftarrow$ image\_features.at(prompt)
\STATE mask $\leftarrow$ SAM.getMask(prompt, frame)
\STATE new\_obj $\leftarrow$ (mask, frame, prompt)
\RETURN new\_obj
\end{algorithmic}
\end{algorithm}

\subsubsection{Using Geometric Features}
\seclabel{SfMSAM}
\algoref{SfMSAM} describes the FIND\_OBJECT implementation when geometric features are included. This algorithm requires the integer parameter $k$, a scene mesh, and a binary variable isRandom. This implementation requires the SfM algorithm to be run beforehand to obtain a scene mesh and camera trajectory. The scene mesh consists of a set of connected triangles describing the geometry of the scene. The camera trajectory is described by the poses of the camera at each frame. This information is sufficient to identify which pixels observe which mesh face. 

For brevity, pixel-triangle matches are included in the mesh object as part of the setup in \algoref{SfMSAM}, mesh.getObj() returns the mesh faces currently known to be part of a known object - this is called the matching set - mesh.getPixelMatches() maps from triangles to pixels in a given frame, mesh.getMeshMatch() maps from pixels to triangles, and mesh.updateObj() adds previously un-seen triangles to a known object.

Given the matching set, a SAM prompt simply needs to be selected from among these pixels. The binary variable isRandom determines the prompt selection methodology. 

Given the object mask from SAM, the mesh object is updated by adding the unseen mesh faces matched to the mask and removing the mesh faces associated to obj\_pixels but not in the SAM mask.

If isRandom is False, visual features are recorded in addition to geometric. In this case, the  $k$-best pixels are selected by visual feature matching as in \secref{SAMonly}. However, only the pixels in the matching set are scored. If isRandom is True, then $k$ pixels are selected at random from among the matching set. In this case, visual features are not tracked and SAM does not need to provide an image encoding.

\figref{ObjectUpdate} shows a visual of the update step for an object in two frames. The figure shows cropped versions of the second (top) and third (bottom) frames of video 1. Each frame has been colored to show the update taking place to the object. A red filter was applied to obj\_pixels; a green filter was applied to obj\_mask, and a blue filter to the pixels not in the obj\_mask. The resulting colors show which pixel-mesh matches are added (green), maintained (yellow), or discarded (purple) as part of the object. As shown, the left wheel was not initially part of the object mesh after the first frame, added in the second frame, persisted in the third while previously out-of-frame pixels are added.

The removal of mesh faces is designed to handle cases with imperfect scene reconstructions so that errors in the reconstruction do not accumulate. More precise reconstructions may not need this removal step.

\begin{algorithm}
\caption{SfM-SAM FIND\_OBJECT} \algolabel{SfMSAM}
\begin{algorithmic}[1]
\REQUIRE frame, obj, SAM, $k$, mesh, isRandom
\STATE obj\_mesh $\leftarrow$ mesh.getObj(obj)
\STATE obj\_pixels $\leftarrow$ mesh.getPixelMatches(frame, obj\_mesh)
\IF{obj\_pixels is empty}
    \RETURN empty
\ENDIF
\IF{isRandom}
    \STATE prompt $\leftarrow$ kRandom(obj\_pixels)
\ELSE
    \STATE image\_features $\leftarrow$ SAM.encodeImage(frame)
    \STATE pixel\_features $\leftarrow$ image\_features.at(obj\_pixels)
    \STATE pixel\_scores $\leftarrow$ Dot(obj.features, pixel\_features)
    \STATE prompt $\leftarrow$ kBest(pixel\_scores)
    \STATE obj.last\_prompt $\leftarrow$ prompt
    \STATE obj.features $\leftarrow$ image\_features.at(prompt)
\ENDIF
\STATE obj\_mask $\leftarrow$ SAM.getMask(prompt, frame)
\STATE mesh\_update $\leftarrow$ mesh.getMeshMatch(obj\_mask, frame)
\STATE mesh.updateObj(obj, mesh\_update)
\STATE new\_obj $\leftarrow$ (mask, frame)
\RETURN new\_obj
\end{algorithmic}
\end{algorithm}

\begin{figure}[h]
    \vspace{3mm}
    \begin{subfigure}{0.45\textwidth}
        \includegraphics[width=\linewidth]{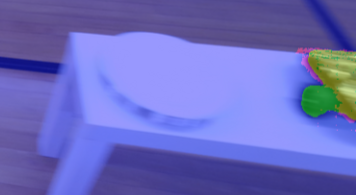}
        \caption{Second Frame Object Update}
    \end{subfigure}
    \begin{subfigure}{0.45\textwidth}
        \includegraphics[width=\linewidth]{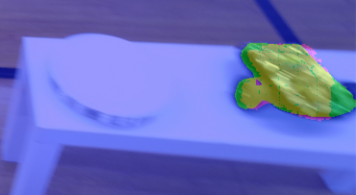}
        \caption{Third Frame Object Update}
    \end{subfigure}
    \caption{(best viewed in color) SfM-SAM object update step; green - pixel-mesh faces added by SAM; red - pixel-mesh faces removed by SAM; yellow - pixel-mesh maintained; blue - not part of object }
    \figlabel{ObjectUpdate}
\end{figure}

\section{Evaluation}
\seclabel{eval}
We evaluate SLP on five videos for time efficiency, accuracy, and tracking loss. SLP is run without any manual labels provided. Objects tracked are objects proposed by SAM.

Time efficiency is determined by both the total run time, and the frame processing rate. Computation is done on a machine with 4 a100 GPUs, 527821928 kB of RAM, and an Intel(R) Xeon(R) Gold 6342 CPU @ 2.80GHz.

To determine accuracy, a ground truth label is needed. Each video is manually labeled and segmented by volunteers to provide ground truth. The manual labels are spot-checked by the authors to ensure quality with ten frames from each video chosen at random. Each selected frame was manually labeled a second time by the authors and quality was assessed by the Intersection over Union (IoU) score between objects.

The IoU score is also used to assess the accuracy of SLP against ground truth. Because no manual labels are supplied to SLP, the objects found by SAM must be matched to the true object mask against which to compare. These matches are determined by finding the SAM mask with the highest IoU score for each manual mask.

A tracking loss occurs when a true label is matched to a different object than what was matched in the previous frame. Such an event indicates that SLP falsely determined that an object was not in frame and a new instance was added. Tracking losses can be considered as a false negative.

False positives would be when SAM recognizes objects that are not manually labeled. However, SAM is designed to recognize objects and component objects independently, while the volunteers were only instructed to label full objects. Therefore, it would not be a fair comparison to evaluate performance with this metric against ground-truth.

The parameters $k$ and $F$ are considered for comparison of methods across different settings. $k$ is the number of pixels used per SAM prompt when identifying known objects in a new frame. $F$ is the number of frames skipped before SAM searches for new objects.

To assess the significance of using geometric features, four variants of SLP are evaluated. Each variant uses a different implementation of FIND\_OBJECT() described in \secref{FindObject} and used in \algoref{SLP}.

\subsubsection{SAM-only-1.0} uses \algoref{SAMonly} with $k\ge0$, described in \secref{SAMonly}, and does not use any geometric features. This is the baseline algorithm.
\subsubsection{SAM-only-2.0} uses \algoref{SAMonly} with $k=0$. This variant scores fewer pixels than SAM-only-1.0.
\subsubsection{SfM-SAM-1.0} uses \algoref{SfMSAM} with isRandom set to False, described in \secref{SfMSAM}. This variant uses mesh faces of a scene to track objects. 
\subsubsection{SfM-SAM-2.0} uses \algoref{SfMSAM} with isRandom set to True. It avoids scoring any pixels and does not track visual features, relying only on geometric features.

\section{Results}
A summary of the data-sets created by each video is shown in \tabref{summary}, each row corresponds to a different video and the column labels describe the following:
\begin{itemize}
    \item $N_f$ - Number of frames in video
    \item $T_{sfm}$ - Time to complete SfM in minutes
    \item $N_m$ - Number of mesh faces
    \item $T_{pxl}$ - Time to complete pixel matching in minutes
    \item $FN_v$ - Number of unlabeled objects in spot check across all frames
    \item $FN_a$ - Number of objects added against spot check across all frames
    \item $\overline{IoU_s}$ - Average IoU across all frames and objects in spot check
\end{itemize}
The last three columns refer to results from the spot check analysis, and the three columns preceding them refer to the SfM setup that is done for the SfM-SAM methods.

\begin{table}[h]
    \centering
    \begin{tabular}{c|c|c|c|c|c|c}
        $N_f$ & $T_{sfm}$& $N_m$ & $T_{pxl}$ & $FN_v$ & $FN_a$ & $\overline{IoU_s}$ \\
        \hline
        281 & 23 & 558440 & 941 & 2 & 1 & 0.940 \\
        \hline
        313 & 24 & 600065 & 1123 & 5 & 1 & 0.925 \\
        \hline
        402 & 57 & 640527 & 1508 & 19 & 1 & 0.709 \\
        \hline
        239 & 15 & 518774 & 692 & 17 & 0 & 0.784 \\
        \hline
        61 & 3 & 380697 & 128 & 13 & 0 & 0.817 \\
        \hline
    \end{tabular}
    \caption{data summary for videos 1-5}
    \tablabel{summary}
\end{table}

\subsection{Spot Check Data}
To calculate the IoU score for each object in each frame, the Author-Labeled Object Mask (ALOM) is compared against each Volunteer-Labeled Object Mask (VLOM), and the VLOM yielding the highest IoU score is taken as the best score for that ALOM. Note that a VLOM might be matched to more than one ALOM if a VLOM overlaps a few ALOM, or if there are more ALOM than VLOM. The authors verified that the former does not happen in any spot-checked frame. $\overline{IoU_s}$ is then the average best-IoU score across all ALOMs and all ten spot-checked frames.

If there are more VLOMS than ALOMS, then the volunteer identified one or more objects the author did not, indicating a false negative on the author's part. However, if there are more ALOMS than VLOMS, then the author identified more objects than the volunteer, which indicates false nagatives made by the volunteer. The total number of such false negatives are tracked by $FN_a$ and $FN_v$ respectively.

Some example results of IoU scores from the spot-check analysis are shown in \figref{spotCheckIOU}. Each line represents an object's IoU score in each spot-checked frame. The frame ID for that video is charted on the x-axis, and the best IoU score on the y-axis. Some lines in \figref{spotCheckIOU} abruptly start or end because not all objects are visible in every frame.

\begin{figure}
    \centering
    \includegraphics[width=\linewidth]{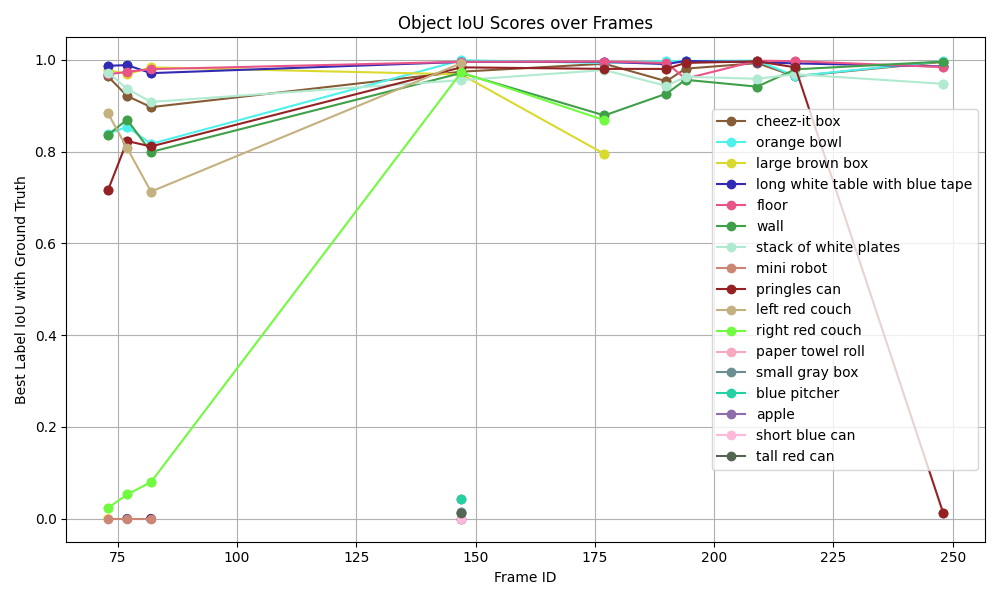}
    \caption{Spot check IoU scores, video 5. }
    \figlabel{spotCheckIOU}
\end{figure}

An example of the discrepancy between author-labeled and volunteer-labeled frames is shown in \figref{AuthorVsVolunteer}. This figure shows frame 73 of video 5 with overlays of the object masks identified by the volunteer (top) and the author (bottom). Each object mask is color-coded by the same pallet shown in the legend of \figref{spotCheckIOU}. The IoU scores for each object can be seen in \figref{spotCheckIOU} with the left-most set of objects.

\begin{figure}[h]
    \centering
    \begin{subfigure}{0.40\textwidth}
        \includegraphics[width=\linewidth]{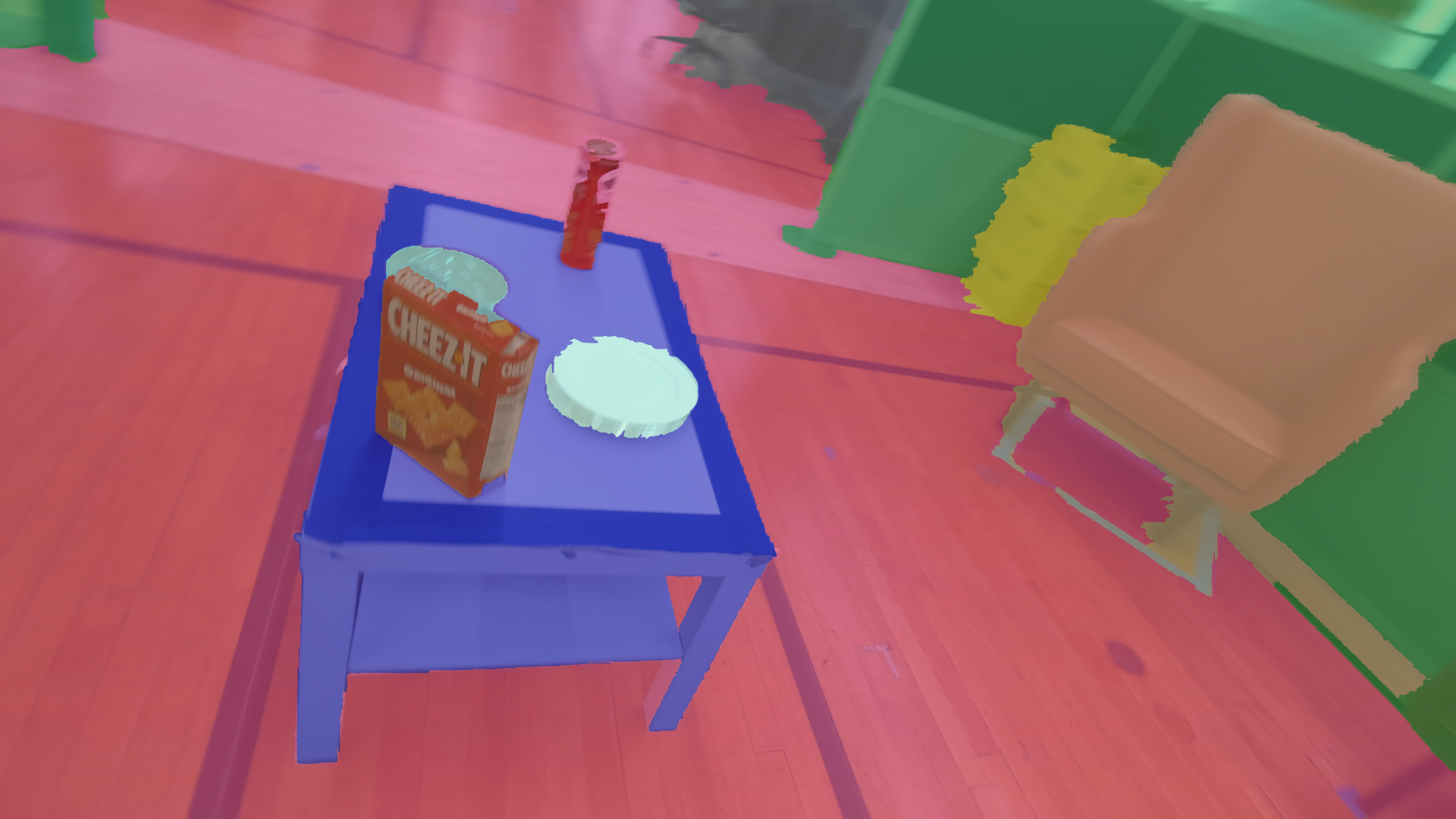}
        \caption{Volunteer Labeled Image}
    \end{subfigure}
    \begin{subfigure}{0.40\textwidth}
        \includegraphics[width=\linewidth]{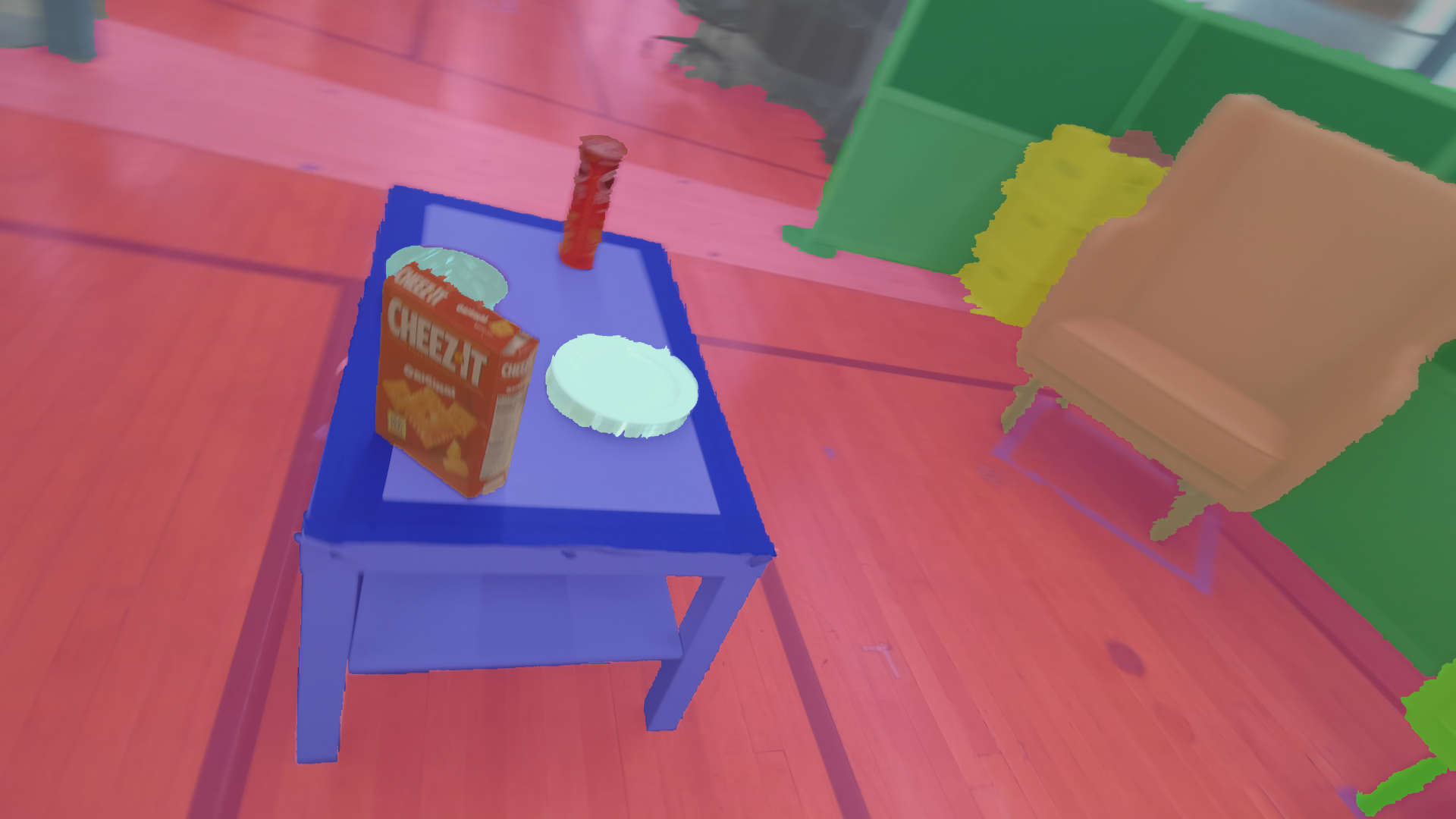}
        \caption{Author Labeled Image}
    \end{subfigure}
    \caption{(best viewed in color) comparison of labeled data (top) to author's spot check (bottom), frame 73, video 5}
    \figlabel{AuthorVsVolunteer}
\end{figure}

The authors labeled one frame every 20min on average, while the volunteers labeled one frame every 10min.

\subsection{Semantic Label Propagation Data}
The performance of SLP is evaluated across different parameter settings in five videos. The best tuned parameters are presented for each of three different performance metrics. 

\tabref{best1} shows the results for the parameter set which gives the fastest computation time. The rows of the table give the implementation indicated by the first column. The next three columns give each of the performance metrics, total compuation time, average IoU, and total number of tracking losses. These metrics are evaluated across all five videos used. The last two columns give the parameter values $k$ and $F$ yielding the fastest computation time, three values of each are tested $k=[1,2,5]$, $F=[0,1,4]$. 

\tabref{best2} and \tabref{best3} are organized in a similar manner, however \tabref{best2} displays the highest average IoU, and \tabref{best3} the lowest number of tracking losses.

\begin{table}
    \vspace{2mm}
  \centering
  \begin{tabular}{|c|c|c|c|c|c|}
    \hline
    Implementaion & Time (min) & IoU & Tracking Loss & $k$ & $F$  \\
    \hline
    SAM-only-1.0 & 260 & 0.396 & 3746 & 2 & 0\\
    \hline
    SAM-only-2.0 & 2109 & 0.323 & 1389 & 1 & 4 \\
    \hline
    SfM-SAM-1.0 & 1741 & 0.188 & 5820 & 5 & 4 \\
    \hline
    SfM-SAM-2.0 & 4116 & 0.222 & 8327 & 5 & 4 \\
    \hline
  \end{tabular}
  \caption{best performing implementation by time}
  \tablabel{best1}
\end{table}

\begin{table}
  \centering
  \begin{tabular}{|c|c|c|c|c|c|}
    \hline
    Implementaion & Time (min) & IoU & Tracking Loss & $k$ & $F$  \\
    \hline
    SAM-only-1.0 & 286 & 0.396 & 3771 & 1 & 1 \\
    \hline
    SAM-only-2.0 & 2188 & 0.396 & 3681 & 1 & 0 \\
    \hline
    SfM-SAM-1.0 & 9200 & 0.253 & 15793 & 1 & 0 \\
    \hline
    SfM-SAM-2.0 & 6765 & 0.274 & 15031 & 2 & 0 \\
    \hline
  \end{tabular}
  \caption{best performing implementation by IoU score}
  \tablabel{best2}
\end{table}

\begin{table}
  \centering
  \begin{tabular}{|c|c|c|c|c|c|}
    \hline
    Implementaion & Time (min) & IoU & Tracking Loss & $k$ & $F$  \\
    \hline
    SAM-only-1.0 & 260 & 0.396 & 3746 & 2 & 0 \\
    \hline
    SAM-only-2.0 & 2110 & 0.323 & 1389 & 1 & 4 \\
    \hline
    SfM-SAM-1.0 & 2786 & 0.192 & 2196 & 2 & 4 \\
    \hline
    SfM-SAM-2.0 & 4116 & 0.222 & 8327 & 5 & 4 \\
    \hline
  \end{tabular}
  \caption{best performing implementation by tracking losses}
  \tablabel{best3}
\end{table}


\section{Discussion}
The fastest implementation of SLP was SAM-only-1.0 at 260min over 1296 frames in all five video data collected. A human equivalent volunteer would have taken 12960min to accomplish the same, while the more meticulous authors would have taken 25,920min. The fastest computation time of SAM-only-1.0 did not suffer on accuracy compared to the other methods. The IoU score for the fastest time was within three decimal place rounding of the fastest IoU score overall, although performing worse than human volunteers. SAM-only-1.0 only had a significant disadvantage on tracking loss.

SAM-only-2.0 and SfM-SAM-1.0 showed improvement tracking losses. It is possible this is due to SAM-only-2.0's use of locality when searching for known objects and the geometric features used in SfM-SAM-1.0. SfM-SAM-2.0 actually had a  performance decrease, possibly because of its random prompt selection leading to sub-optimal outcomes. 

\section{Conclusion}
This work presents a system named Semantic Label Propagation for automatic labeling of video data suitable for training a semantic segmentation model. The system tracks visual and geometric features of objects across frames to track segments which are part of previously identified objects. The trade-off in computational effort versus manual effort means that this technique can be incorporated into a tool to speed up human annotations by allowing annotators to make a few corrections to an annotated video, rather than manually labeling each frame; speeding up their overall effort.





\section*{ACKNOWLEDGMENT}
This work has taken place in the Living with Robots Laboratory (LWR) at UT Austin. LWR research is supported in part by NSF (NRT-2125858 and GCR-2219236), Cisco Research, and Army Futures Command.
\clearpage 
\printbibliography


\end{document}